\pdfoutput=1

\documentclass[runningheads]{llncs}
\usepackage{llncsdoc}

\usepackage{amsmath}
\usepackage{latexsym}
\usepackage{epsfig}
\usepackage{verbatim}
\usepackage{color}
\usepackage{qtree} 

\title{A Compositional Distributional Semantics, \\ Two Concrete Constructions,  \\  and some Experimental Evaluations}
\author{Mehrnoosh Sadrzadeh\thanks{
Support by EPSRC (grant EP$/$F042728$/$1) is gratefully acknowledged.} and Edward Grefenstette}
\institute{Department of Computer Science, University of Oxford, UK. \\
\ \\
\email{mehrnoosh.sadrzadeh@cs.ox.ac.uk} \qquad \email{edward.grefenstette@cs.ox.ac.uk} }

\titlerunning{A compositional distributional model of meaning}

\authorrunning{M. Sadrzadeh and E. Grefenstette}

\begin{document}
\maketitle

\begin{abstract}
We provide an overview of the hybrid compositional distributional model of meaning, developed in~\cite{Coeckeetal},  which is based on the categorical methods also applied  to the analysis of information flow in quantum protocols.  The mathematical setting stipulates that the meaning of a sentence is a linear function of the tensor products of the meanings of its words. We provide concrete constructions for this definition and present techniques to build vector spaces for meaning vectors of words, as well as that of sentences. The applicability of these methods is demonstrated via a toy vector space as well as real data from the British National Corpus and two disambiguation experiments.


\end{abstract}

%
%


%








\section{Introduction}

Words are the building blocks of sentences, yet the meaning of a sentence goes well beyond the meanings of its words.  Indeed, while we do have dictionaries for words, we don't seem to need them to  infer meanings of  sentences. But where human beings seem comfortable doing this, machines fail to deliver. Automated search engines that  perform well when queried by single words,  fail to shine  when it comes to  search for meanings of phrases and sentences. Discovering the process of meaning assignment in natural language is among the most challenging as well as  foundational questions of linguistics and computer science. The findings thereof will increase our understanding of cognition and intelligence and will also assist in applications to automating language-related tasks such as  document search.  

To date, the compositional type-logical~\cite{Montague,Lambek} and the distributional vector space models~\cite{Schutze,Firth} have provided  two complementary partial solutions to the question.  The logical approach is based on classic ideas from mathematical logic, mainly Frege's principle  that meaning of a sentence can be derived from the relations of the words in it. The distributional model is more recent, it can be related to Wittgenstein's philosophy of `meaning as use', whereby  meanings of  words can be determined from their context.  The logical models have been the champions of the theory side, but in practice  their distributional rivals have provided the best predictions.

In a cross-disciplinary approach,~\cite{Coeckeetal} used techniques from logic, category theory, and quantum information to develop a compositional distributional semantics that brought the above two models  together. They developed a hybrid categorical model which paired contextual meaning with grammatical form and defined meaning of a string of words to be a function of the tensor product of the meanings of its words. As a result, meanings of sentences became vectors  which lived in the same vector space and it became possible to measure their synonymity the same way lexical synonymity was measured in the distributional models.  This sentence space  was taken to be an abstract space and it was only shown how to instantiate it for the  truth-functional meaning.  Later~\cite{Grefenetal} introduced a concrete construction using structured vector spaces and exemplified the application of logical methods, albeit only a toy vector space. In this paper we report on this and on a second construction which uses plain vector spaces. We also review results on implementing and evaluating the setting on real large scale data from the British National Corpus and two disambiguation experiments~\cite{GrefenSadr}.

\section{Sketching the problem and a hybrid solution}

To compute the meaning of a sentence consisting of $n$ words, meanings of these  words must \emph{interact} with one another.  In the logical models of meaning, this further interaction  is  represented in a  function computed  from the grammatical structure of the sentence, but meanings of words are empty entities. The grammatical structure is usually depicted as a parse-tree, for instance the parse-tree of the transitive sentence `dogs chase cats' is as follows:

\medskip

\Tree [.$\textrm{chase}(\textrm{dogs},\textrm{cats})$ $\textrm{dogs}$ [.$\quad \lambda x.\textrm{chase}(x,|\textrm{cats})$   $\textrm{cats}$ $\quad\lambda yx.\textrm{chase}(x,y)$ ] ]

\noindent
The function  corresponding to this tree is based on a relational reading of the meaning of the verb `chase', which makes the subject and the object interact with each other via the relation of \emph{chasing}. This methodology is used to  translate sentences of natural language into logical formulae, then use computer-aided automation tools to reason about them~\cite{Alshawi}. The  major  drawback is that the result can only deal with truth or falsity as the meaning of a sentence and does poorly on lexical semantics, hence do not perform well on language tasks such as search.

The vector space model, on the other hand,  dismisses the further interaction and is solely based  on lexical semantics. These are obtained in an operational way, best described by a frequently cited quotation due to Firth~\cite{Firth} that ``You shall know a word by the company it keeps.". For instance, beer and sherry are both drinks,  alcoholic, and often make you drunk. These facts are reflected in the text:  words `beer' and `sherry' occur close to `drink',  `alcoholic' and `drunk'. Hence meanings of words  can be encoded as   vectors in a highly dimensional space of context words. The raw weight in each base is related to the number of times the word has appeared close (in an $n$-word window) to that base.   This setting offers geometric means to reason about meaning similarity, e.g.~via the cosine of the angle between the vectors.   
Computational models along
these lines have been built using large vector spaces (tens of
thousands of basis vectors) and large bodies of text (up
to a billion words)~\cite{Curran}.  
These models have responded well to  language processing tasks such as word sense discrimination, thesaurus construction,  and document retrieval~\cite{Grefenstette,Schutze}. 
Their  major drawback is their non-compositional nature: they ignore the grammatical structure and  logical words,   hence cannot  compute (in the same efficient way that they do for words) meanings of phrases and sentences.

The key idea behind the approach of~\cite{Coeckeetal} is  to import the compositional element of the logical approaches into the vector space models by  making  the grammar of the sentence  \emph{act} on, hence relate,  its word vectors. The trouble is that it does not make so much sense to `make a parse tree act on vectors'. Some higher order mathematics, in this case  category theory, is needed to encode the grammar of a sentence into a morphism compatible with vector spaces\footnote{A similar passage had to be made in other type-logics to turn the parse-trees into lambda terms, compatible with sets and relations.}.  These morphisms turn out to be the grammatical reductions of a type-logic called a Lambek pregroup~\cite{Lambek}. Pregroups and vector spaces both have a \emph{compact} categorical structural. The   grammatical morphism of a pregroup can be transformed into a linear map that acts on vectors.  Meanings of sentences become vectors whose angles reflect similarity. Hence, at least theoretically, one should be able to build sentence vectors and compare their synonymity, in exactly the same way as measuring synonymity for  words. 

The pragmatic interpretation of this abstract idea is as follows.  In the vector space models,  one has a meaning vector for each word, $\overrightarrow{\text{dogs}}, \overrightarrow{\text{chase}}, \overrightarrow{\text{cats}}$. The  logical recipe tells us to \emph{apply} the meaning of verb to the meanings of subject and object. But how can a vector \emph{apply} to other vectors? If we  strip the vectors  off the extra information  provided in their basis and look at them as mere sets of weights, then we can apply them to each other by taking their point-wise sum or product. But these operations are commutative, whereas meaning is not. Hence this will equalize meaning of any combination of words, even with the non-grammatical combinations such as `dogs cats chase'.  The proposed solution above implies that one needs to  have different levels of meaning for words with different functionalities. This is similar to the logical models whereby verbs are relations and  nouns are atomic sets. So verb vectors should be built differently from noun vectors, for instance as matrices that relate and act on the atomic noun vectors. The general information, as to which words should be matrices and which atomic vectors,  is in fact encoded in the type-logical representation of the grammar. That is why the  grammatical structure of the sentence is a good candidate for the process that relates  its word vectors.

In a nutshell, pregroup types are either atomic or compound. Atomic types can be simple (\emph{e.g.}~$n$ for noun phrases, $s$ for statements) or left/right superscripted---referred to as adjoint types (\emph{e.g.}~$n^r$ and $n^l$). An example of a compound type is that of a verb  $n^r s n^l$.  The superscripted types  express that the verb is a relation with two arguments of type $n$, which have to occur to the \underline{\emph{r}}ight and to the \underline{\emph{l}}eft of it, and that it outputs an argument of the type $s$. A transitive sentence is typed  as shown below.

\vspace{-2mm}
\begin{center}
\begin{minipage}{3cm} 
\hspace{-2cm}\begin{picture}(50,50)(150,150)
\put(200,185){dogs}
\put(208,170){$n$}
\put(230,185){chase}
\put(228,170){$n^r$}
\put(240,170){$s$}
\put(248,170){$n^l$}
\put(260,185){cats.}
\put(270,170){$n$}
\end{picture}

\vspace{-0.5cm}
\hspace{-0.15cm}{\epsfig{figure=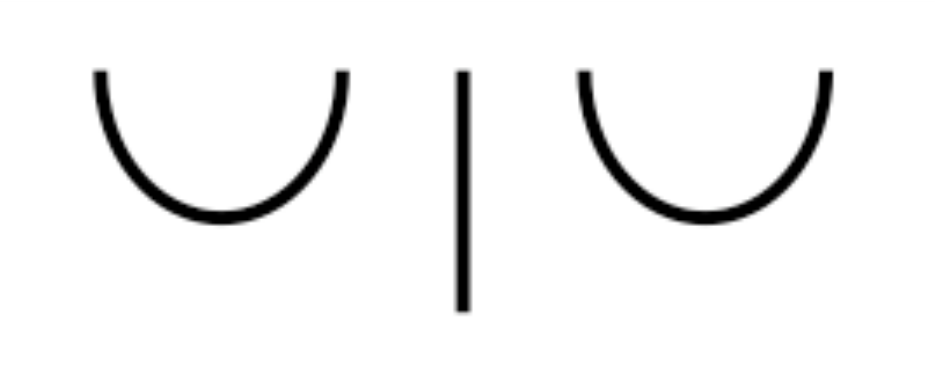,width=80pt}} 
\end{minipage}
\end{center}

\vspace{-2mm}
\noindent
Here, the verb interacts with the subject and object via the underlying wire cups, then produces a sentence via the outgoing line.  These interactions happen in real time. The type-logical analysis assigns type $n$ to `dogs' and `cats', for a noun phrase, and the type $n^r s n^l$ to `chase' for a verb, the superscripted types $n^r$ and $n^l$ express the fact that the verb is a function with two arguments of type $n$, which have to occur to the \emph{r}ight and \emph{l}eft of it. The reduction computation is $nn^rsn^l \leq 1s1 = s$, each type $n$ cancels out with its right \emph{adjoint} $n^r$ from the right, i.e. $nn^r \leq 1$ and its left adjoint $n^l$ from the left, i.e. $n^ln\leq 1$, and 1 is the unit of concatenation $1n=n1=n$. The algebra  advocates   a linear method of parsing: a   sentence is analyzed as it is heard, i.e.~word by  word,  rather than by first buffering the entire string then  re-adjusting it as necessary on a tree. It's been argued  that  the  brain works in this one-dimensional linear (rather than two-dimensional tree) manner~\cite{Lambek}. 
%



According to~\cite{Coeckeetal} and based on a general completeness theorem between compact categories, wire diagrams, and vector spaces,  meaning of sentences can be canonically reduced to linear algebraic formulae, for example the following is the meaning vector of our transitive sentence:
\[
\overrightarrow{\text{dogs} \ \text{chase} \ \text{cats}}  =  
(f)\!\left(\overrightarrow{\text{dogs}} \otimes \overrightarrow{\text{chase}} \otimes \overrightarrow{\text{cats}}\right)
\]  
Here $f$ is the  linear map  that encodes the grammatical structure. The categorical morphism corresponding to it is denoted by the tensor product of 3 components: $\epsilon_V \otimes 1_S \otimes \epsilon_W$, where $V$ and $W$ are subject and object spaces, $S$ is the sentence space,  the $\epsilon$'s are the cups, and $1_S$ is the straight line in the diagram.   The cups stand for taking inner products, which when done with the basis vectors imitate substitution. The straight line stands for the identity map that does nothing. By the rules of the category,  the above equation  reduces to  the following  linear algebraic formula with lower dimensions, hence the dimensional explosion problem for tensor products is avoided:

\[
{\sum_{itj} C^{\text{chase}}_{itj} \langle \overrightarrow{\text{dogs}} \mid  \overrightarrow{v_i}\rangle  \overrightarrow{s_t} 
\langle  \overrightarrow{w_j}\mid \overrightarrow{\text{cats}}\rangle\in S} 
\]


\noindent
In the above equation,
$\overrightarrow{v_i}, \overrightarrow{w_j}$ are basis vectors of $V$ and $W$. The meaning of the verb becomes a superposition, represented as a linear map. The inner product  $\langle \overrightarrow{\text{dogs}}\!\! \mid \!\! \overrightarrow{v_i}\rangle $ substitutes the weights of $\overrightarrow{\text{dogs}}$ into the first argument place of the verb (similarly for object and second argument place) and results in producing a vector for the meaning of the sentence.  These vectors live in sentence spaces $S$, for which  $\overrightarrow{s_t}$ is a base vector.   The  degree of synonymity of sentences is  obtained by taking the cosine measure of  their vectors.    $S$ is an abstract space, it needs to be instantiated to provide concrete meanings and synonymity  measures. For instance, a truth-theoretic model is obtained by taking the sentence space  $S$  to be the 2-dimensional space with basis vector true $\mid\!\!1\rangle$ and false $\mid\!\!0\rangle$. This is done by  using the weighting factor $C^{\text{chase}}_{itj}$ to define a model-theoretic meaning for the verb as follows:
\[
C^{\text{chase}}_{itj} \overrightarrow{s_t} = \begin{cases} \mid\!\!1\rangle &
{chase} ({v_i},{w_j})  =  \text{true}\,,\\  \mid\!\!0\rangle & o.w.\end{cases}
\]
The definition of our meaning map ensures that this value propagates
to the meaning of the whole sentence. So $chase(\text{dogs}, \text{cats})$ becomes true whenever `dogs chase cats' is true and false otherwise. 

\section{Two Concrete Constructions for Sentence Spaces} 
The above construction is based on the assumptions that $\overrightarrow{\text{dogs}}$ is a base of  $V$ and that $\overrightarrow{\text{cats}}$ is a base of  $W$. In other words, we assume that $V$ is the vector space spanned by the set of all men and $W$ is the vector space spanned by the set of all women. This is not the usual construction in the distributional models. In what follows we present two concrete constructions for these, which will then yield a construction for the sentence space. In both of these approaches $V$ and $W$ will be the same vector space, which we will denote by $N$.

\subsection{Structured Vector Spaces and a Toy Corpus}

We take  $N$ to be  a \emph{structured vector
space}, as in~\cite{Grefenstette}. The bases of $N$ are
annotated by 
`properties' obtained by combining dependency relations with nouns, verbs and adjectives. 
For example, basis vectors might be associated with properties such as ``arg-fluffy'', denoting the argument of the adjective fluffy, ``subj-chase'' denoting the subject of the verb chase, ``obj-buy'' denoting the object of the verb buy, and so on. We construct the vector for a noun by counting how many times in the corpus  a word has been the argument of `fluffy', the subject of `chase', the object of `buy', and so on. 

For transitive sentences, we take the sentence space $S$ to be  $N \otimes N$,  so
its bases are of the form $\overrightarrow{s_t} =
{(\overrightarrow{n_i}, \overrightarrow {n_j})}$.  The intuition is that, for a transitive verb, the meaning of a sentence is determined by the meaning of the verb together with its subject and object.
The verb vectors
$C^{\text{verb}}_{itj}{(\overrightarrow{n_i}, \overrightarrow
{n_j})}$ are built by  counting how
many times a word that is  $n_i$ (e.g. has the property of being fluffy) has been subject of the verb and
a word that is $n_j$ (e.g. has the property that it's bought) has been its object, where the counts are moderated by the extent to which the subject and object exemplify each property (e.g. \emph{how fluffy} the subject is). To give a rough paraphrase of the intuition behind this approach, the meaning of ``dog chases cat" is given by: the extent to which a dog is fluffy and a cat is something that is bought (for the $N \otimes N$ property pair ``arg-fluffy" and ``obj-buy"), and the extent to which fluffy things {\em chase} things that are bought (accounting for the meaning of the verb for this particular property pair); plus the extent to which  a dog is something that runs and a cat is something that is cute (for the $N \otimes N$ pair ``subj-run" and ``arg-cute"), and the extent to which things that run {\em chase} things that are cute (accounting for the meaning of the verb for this particular property pair); and so on for all noun property pairs.

For sentences with intransitive verbs, the sentence
space suffices to be just $N$.  To compare the meaning of a transitive
sentence with an intransitive one, we embed the meaning of the latter
from $N$ into the former $N \otimes N$, by taking
$\overrightarrow{\varepsilon_n}$ (the `object' of an intransitive verb) to be $\sum_i{\overrightarrow{n_i}}$,
i.e. the superposition of all basis vectors of $N$. A similar method is used while dealing with sentences with ditransitive verbs, where the sentence space will be $N \otimes N \otimes N$, since these verbs have three arguments. Transitive and intransitive sentences are then embedded in this bigger space, using the same embedding  described above.

Adjectives are dealt with in
a similar way. We give them the syntactic type $nn^l$ and
build their vectors in $N \otimes N$. The syntactic reduction $nn^l n
\to n$ associated with applying an adjective to a noun gives us the
map $1_N \otimes \epsilon_N$ by which we semantically compose an
adjective with a noun, as follows: \begin{displaymath}
	\overrightarrow{\text{adjective noun}} = (1_N \otimes \epsilon_N)(\overrightarrow{\text{adj}} \otimes \overrightarrow{\text{noun}}) = \sum_{ij}{C^{\text{adj}}_{ij}\overrightarrow{n_i} \langle \overrightarrow{n_j} \mid \overrightarrow{\text{noun}} \rangle}
\end{displaymath}
We can view the $C^{\text{adj}}_{ij}$ counts as determining what sorts of properties the arguments of a particular adjective
typically have (e.g. arg-red, arg-colourful for the adjective
``red'').

As an example, consider a hypothetical  vector space with bases `arg-fluffy', `arg-ferocious', `obj-buys', `arg-shrewd', `arg-valuable', with vectors for `bankers', `cats', `dogs', `stock', and `kittens'.  

\begin{center}
	\begin{tabular}{|c|l|ccccc|}
\hline
		&& bankers & cats & dogs & stock & kittens\\
		\hline\hline
		1& arg-fluffy 		& 0 & 7 & 3 & 0 & 2\\
		2& arg-ferocious 	& 4 & 1 & 6 & 0 & 0\\
		3& obj-buys 		& 0 & 4 & 2 & 7 & 0\\
		4& arg-shrewd 		& 6 & 3 & 1 & 0 & 1\\
		5& arg-valuable 	& 0 & 1 & 2 & 8 & 0\\
		\hline
	\end{tabular}
\end{center}
Since in the method proposed above, $C^{\text{verb}}_{itj}=0$ if $\overrightarrow{s_t} \neq (\overrightarrow{n_i},\overrightarrow{n_j})$, we can simplify the weight matrices for transitive verbs to two dimensional $C^{\text{verb}}_{ij}$ matrices as shown below, where $C^{\text{verb}}_{ij}$ corresponds to the number of times the verb has a subject with attribute $n_i$ and an object with attribute $n_j$. For example, the matrix below encodes the fact that something ferocious ($i=2$) chases something fluffy ($j=1$) seven times in the hypothetical corpus from which we might have obtained these distributions.
\begin{displaymath}
	C^{\textrm{chase}} = \left[
	\begin{tabular}{ccccc}
	1 \quad & 0 \quad& 0 \quad& 0 \quad& 0 \\
	7 \quad& 1 \quad& 2 \quad& 3 \quad& 1 \\
	0 \quad& 0 \quad& 0 \quad & 0 \quad& 0   \\
	2 \quad& 0 \quad& 1 \quad& 0 \quad& 1  \\
	1 \quad& 0 \quad& 0 \quad& 0 \quad & 0   \\
	\end{tabular}
	\right]
\end{displaymath}	
Once we have built matrices for verbs, we are able to  follow the categorical procedure and automatically build vectors for sentences, then   perform sentence comparisons. The comparison  is done in the same way as for lexical semantics, i.e. by taking the inner product of the vectors of two sentences  and normalizing it by the product of their lengths. For example the following shows a high similarity
\[
cos(\overrightarrow{\text{dogs chase cats}},\overrightarrow{\text{dogs pursue kittens}}) = { \langle \overrightarrow{\text{dogs chase cats}} \mid \overrightarrow{\text{dogs pursue kittens}} \rangle  
\over
\mid \overrightarrow{\text{dogs chase cats}} \mid \times \mid \overrightarrow{\text{dogs pursue kittens}}\mid
} = 
\]
\[
{ \left\langle \left(\sum_{itj}{C^{\text{chase}}_{itj} \langle \overrightarrow{\text{dogs}} \mid \overrightarrow{n_i}\rangle  \overrightarrow{s_t} \langle \overrightarrow{n_j}\mid \overrightarrow{\text{cats}}\rangle}\right) \right| \left. \left(\sum_{itj}{C^{\text{pursue}}_{itj} \langle \overrightarrow{\text{dogs}} \mid \overrightarrow{n_i}\rangle  \overrightarrow{s_t} \langle \overrightarrow{n_j}\mid \overrightarrow{\text{kittens}}\rangle} \right) \right\rangle
\over
\mid \overrightarrow{\text{dogs chase cats}} \mid \times \mid \overrightarrow{\text{dogs pursue kittens}}\mid
}
 \]
 \[
  = {\sum_{itj}{C^{\text{chase}}_{itj}C^{\text{pursue}}_{itj} \langle \overrightarrow{\text{dogs}} \mid \overrightarrow{n_i}\rangle \langle \overrightarrow{\text{dogs}} \mid \overrightarrow{n_i}\rangle \langle \overrightarrow{n_j}\mid \overrightarrow{\text{cats}}\rangle \langle \overrightarrow{n_j}\mid \overrightarrow{\text{kittens}}\rangle} 
  \over
\mid \overrightarrow{\text{dogs chase cats}} \mid \times \mid \overrightarrow{\text{dogs pursue kittens}}\mid
} = 0.979
\]
A similar computation will provide us with the following, demonstrating a low similarity 
\[
cos(\langle \overrightarrow{\text{dogs chase cats}}\mid \overrightarrow{\text{bankers sell stock}} \rangle)  = 0.042
\]

The construction for adjective matrices are similar:  we stipulate the $C^{\text{adj}}_{ij}$ matrices by hand and  eliminate all cases where $i \neq j$ since $C_{ij} = 0$, hence these become one dimensional matrices. Here is an example
\[
C^{\text{fluffy}}  = [9\ 3\ 4\ 2\ 2] 
\]
Vectors for `adjective noun' clauses are computed similarly and are used to compute the following similarity measures:
\begin{eqnarray*}
cosine ( \overrightarrow{\text{fluffy \ dog}} ,  \overrightarrow{\text{shrewd \ banker}} ) &=& 0.389\\
cosine ( \overrightarrow{\text{fluffy \ cat}} ,  \overrightarrow{\text{valuable \ stock}}) &=& 0.184
\end{eqnarray*}
These calculations carry over to sentences which contain the `adjective noun' clauses. For instance,  we obtain an  even lower similarity measure between the following sentences:
\[cosine( \overrightarrow{\text{fluffy dogs chase fluffy cats}} , \overrightarrow{\text{shrewd bankers sell valuable stock}} ) = 0.016
\]
Other constructs such as prepositional phrases and adverbs are treated similarly, see~\cite{Grefenetal}.

\subsection{Plain Vector Spaces and the BNC}
The above concrete example is fine grained, but involves complex constructions which are time and space costly when implemented. To be able to evaluate the setting against real large scale data, we simplified it by taking  $N$ to be a plain vector spaces whose bases are words, without annotations. The weighting factor  $C^{\text{verb}}_{ij}$ is determined in the same as above, but this time by just counting co-occurence rather than being arguments of syntactic roles. More precisely, this weight is determined by the number of times  the subjects of the verb have co-occured with the base $\overrightarrow{n}_i$. In the previous construction we went  beyond co-occurence  and required that the subject (similarly for the object) should be in a certain relation with the verb, for instance if $\overrightarrow{n}_i$ was `arg-fluffly', the subject had to be an argument of fluffy, where as here we instead have $\overrightarrow{n}_i$ = `fluffy', and the subject has to co-occure with `fluffy' rather than being directly modified by it.

The  procedure for computing these weights for the case of  transitive sentences is as follows: first browse the corpus to find all occurrences of the  verb in question, suppose it has occurred as a transitive verb in $k$ sentences.  For each sentence determine the subject and the object of the verb. Build vectors for each of these using the usual distributional method. Multiply their weights on all permutations of their  coordinates and then take the sum of each such multiplication across each of the $k$ sentences. Linear algebraically, this is just the sum of the Kronecker products of the vectors of subjects and objects: 
\[\overrightarrow{\text{verb}} \quad =\quad
\sum_{k} \left(\overrightarrow{\text{sub}} \otimes \overrightarrow{\text{obj}}\right)_k
\]
Recall that given a vector space $A$ with basis $\{\overrightarrow{n_i}\}_i$,   the Kronecker product of two vectors $\overrightarrow{v} = \sum_i{c^{a}_i \overrightarrow{n_i}}$ and $\overrightarrow{w} = \sum_i{c^{b}_i \overrightarrow{n_i}}$  is defined as follows:
\[
\overrightarrow{v} \otimes \overrightarrow{w} = \sum_{ij}{c^a_i c^b_j\,  (\overrightarrow{n_i}\otimes\overrightarrow{n_j}})
\]

As an example, we worked with the British National Corpus (BNC) which has  about 6 million sentences. We built noun vectors and computed matrices for intransitive verbs, transitive verbs,  and adjectives. For instance, consider $N$ to be the space with four basis vectors `far', `room', `scientific', and `elect'; the (TF/IDF) values for vectors of the four nouns `table', `map', `result', and `location'  are shown below.
\begin{table}
	\begin{center}
		\begin{tabular}{|c|c|c|c|c|c|}
		\hline
		$\mathbf{i}$ & $\overrightarrow{\mathbf{n_i}}$ &  table & map & result & location \\
		\hline
		\hline

		1&far &6.6 & 5.6 &7& 5.9\\

		2&room &  27 & 7.4 &0.99& 7.3\\

		3&scientific& 0& 5.4& 13& 6.1\\

		4&elect & 0  & 0&4.2&0\\
		\hline
		\end{tabular}
		\label{samplenoun}
		\caption{Sample noun vectors from the BNC.}
	\end{center}
	\end{table}

\noindent
A section of the matrix of the transitive verb `show' is represented below.

\begin{table}
	\begin{center}
		\begin{tabular}{|c|c|c|c|c|}
		\hline

		&far & room & scientific & elect\\
		\hline

		far &\fbox{79.24} &47.41&119.96&27.72\\
		room& 232.66 & 80.75&396.14& 113.2\\
		scientific &32.94&31.86 &32.94&0\\
		elect&0&0&0&0\\
		\hline
		\end{tabular}
		\end{center}
		\label{sampleverb}
		\caption{Sample verb matix from the BNC.}
\end{table}
 
 \medskip
 \noindent
As a sample computation, suppose the verb `show' only appears in two sentences in the corpuse: `the map showed the location' and `the table showed the result'. The weight  $c_{12}$ for  the base \emph{i.e.}~$(\overrightarrow{\text{far}}, \overrightarrow{\text{far}})$ is computed by multiplying weights of `table' and `result' on $\overrightarrow{\text{far}}$,  \emph{i.e.}~$6.6 \times 7$, multiplying weights of `map' and `location' on $\overrightarrow{\text{far}}$, \emph{i.e.}~$5.6 \times 5.9$ then adding these $46.2 + 33.04$ and obtaining the total weight $79.24$. 

The computations for building vectors for sentences and other phrases are the same as in the case for structured vector spaces.  The matrix of a transitive verb has 2 dimensions since it takes as input two arguments. The same method is applied to build matrices for ditransitive verbs, which will have 3 dimensions, and intransitive verbs, as well as adjectives and adverbs, which will be of 1 dimension each.  

\section{Evaluation and Experiments}
We evaluated our second concrete method on a disambiguation task and performed two experiments~\cite{GrefenSadr}. The general idea behind this disambiguation task is that some verbs have different meanings  and the context in which they appear is used to disambiguate them. For instance the verb `show' can mean `express' in the context `the table showed the result' or it can mean 'picture', in the context `the map showed the location'. Hence if we build meaning vectors for these sentences compositionally, the degrees of synonymity of the sentences can be used to disambiguate the meaning of the verb in that sentence. Suppose a verb has two meanings and it has occurred in two sentences.  Then if in both of these sentences it has its meaning number 1, the two sentences will have a high degree of synonymity, whereas if in one sentence the verb has its meaning number 1 and in the other its meaning number 2, the sentences will have a lower degree of synonymity. For instance,  `the table showed the result' and `the table expressed the result',  have a hight degree of synonymity and similarly for   `the map showed the location' and `the map pictured the location'. This degree decreases for  the two sentences `the table showed the result' and `the table pictured the result'. We used our second concrete construction to implement this task.

The data set for our first experiment was  developed by~\cite{Lapata} and  had 120 sentence pairs. These were all intransitive sentences.  We compared the results of our method with composition operations implemented by~\cite{Lapata}, these included addition, multiplication, and a combination of two using weights. The best results were obtained by the multiplication operator. Our method provided slightly better results. However, the context provided by intransitive sentences is just one word, hence the results do not showcase the compositional abilities of our method. In particular, in such a small context, our method and the multiplication method became very similar, hence the similarity of results did not surprise us. There is nevertheless two  major differences:  our method respects the grammatical structure of the sentences (whereas the multiplication operation does not) and in our method the vector of the verb is computed differently from the vectors of the nouns: as a relation and via a second order construction.

For the second experiment, we developed a data set of transitive sentences. We first picked 20 transitive verbs from the most occurring verbs of the BNC, each verb has at least two different non-overlapping meanings. These were retrieved using the JCN (Jiang Conrath) information content synonymity measure of  WordNet. The above example for `show' and its two meanings `express' and `picture' is one such example. For each such verb, e.g. `show', we retrieved 10 sentences which contained  them (as verbs) from the BNC. An example of such a sentence is `the table showed the result'. We then substituted in each sentence each of the two meanings of the verb, for instance `the table expressed the result' and `the table pictured the result'. This provided us with 400 pairs of sentences and we used the plain method described above to build vectors for each sentence and compute the cosine of each pair. A sample of these pairs is provided below. 

\begin{table}[h]
\begin{center}
 \begin{tabular}{|c|c|c|}
 \hline
 &Sentence 1 & Sentence 2\\
 \hline
 \hline
 
1& table show result & table express  result  \\
\hline
2&  table show result & table picture result\\
\hline
3& map show location & map picture location\\
 \hline
 4& map show location & map express location\\
 \hline
5&  child show  interest  & child picture  interest \\
 \hline
 6&  child show  interest  & child express  interest \\
\hline
\end{tabular}
\end{center}
\label{sampledataset}
\caption{Sample sentence pairs from the second experiment dataset.}
\end{table}

In order to  judge the performance of our method, we followed guidelines from~\cite{Lapata}.  We distributed our data set among  25 volunteers who were asked to  rank each pair based on how similar they thought they were. The ranking was between 1 and 7, where 1 was almost dissimilar and 7 almost identical. Each pair was also given a HIGH or LOW classification by us. The correlation of the model's similarity judgements with the human judgements was calculated using Spearman's $\rho$, a metric which is deemed to be more scrupulous and ultimately that by which models should be ranked. It is assumed that inter-annotator agreement provides the theoretical maximum $\rho$ for any model for this experiment, and that taking the cosine measure of the verb vectors while ignoring the noun was taken as the baseline.

The results for the models evaluated against the both datasets are presented below. The additive and multiplicative  operations are applications of vector addition and  multiplication; Kintsch is a combination of the two, obtained by multiplying the word vectors by certain weighting constants and then  adding them, for details please see~\cite{Lapata}. The \emph{Baseline} is from a non-compositional approach, obtained by only comparing vectors of verbs of the sentences and ignoring their subjects and objects. The \emph{UpperBound} is the summary of the human ratings, also known as inter-annotator agreement.

\begin{table}[h]
\begin{center}
\begin{tabular}{|lll|l|}
\hline
Model & High & \quad Low & $\quad \rho$\\
\hline
\hline
Baseline & 0.27 & \quad 0.26 & \quad 0.08\\
\hline
\hline
Add & 0.59 & \quad 0.59 & \quad 0.04\\
Kintsch & 0.47 & \quad 0.45 & \quad 0.09\\
Multiply & 0.42 & \quad 0.28 & \quad 0.17 \\
\textbf{Categorical} & \textbf{0.84} & \quad\textbf{0.79} & \quad\textbf{0.17}\\
\hline
\hline
UpperBound & 4.94 & \quad 3.25 & \quad 0.40 \\
\hline
\end{tabular}\hspace{1cm}
\begin{tabular}{|lll|c|}
\hline
Model & High & \quad Low & \quad $\rho$\\
\hline
\hline
Baseline & 0.47 & \quad 0.44 & \quad 0.16\\
\hline
\hline
Add & 0.90 & \quad 0.90 & \quad 0.05\\
Multiply & 0.67 & \quad 0.59 & \quad 0.17 \\
\textbf{Categorical} & \textbf{0.73} & \quad\textbf{0.72} & \quad \textbf{0.21}\\
\hline
\hline
UpperBound & 4.80 & \quad 2.49 & \quad 0.62 \\
\hline
\end{tabular}
\end{center}
\label{results}
\caption{Results of the 1st and 2nd compositional disambiguation experiments.}
\end{table}

According to the literature (e.g. see~\cite{Lapata}), the main measure of success is demonstrated by the $\rho$ column. By this measure in the second experiment our  method outperforms the other two  with a much better margin than that in the first experiment. The High (similarly Low) columns are the average score that High (Low) similarity sentences (as decided by us) get by the program. These are not very indicative, as the difference between high mean and the low mean of the categorical model is much smaller than that of the both the baseline model and multiplicative model, despite better alignment with annotator judgements.

The data set of the first experiment  has a very simple syntactic structure where the context around the verb is just its subject.  As a result,  in practice the categorical method becomes very similar to the multiplicative one and the similar outcomes should not surprise us. The second experiment, on the other hand, has more syntactic structure, thereby our categorical shows an increase in alignment with human judgements.  Finally, the increase of $\rho$ from the first experiment to the second reflects the compositionality  of our model:  its performance increases with the increase in syntactic complexity.  Based on this, we would like to believe  that more complex datasets and experiments which for example include adjectives and adverbs  shall lead to even better results.

\section{Conclusion and Future Work}
We have provided a brief  overview of the categorical compositional distributional model of meaning as developed in~\cite{Coeckeetal}. This combines the logical and vector space models using the setting of compact closed categories and their diagrammatic toolkit and based on ideas presented in~\cite{ClarkPulman} on the use of tensor product as a meaning composition operator. We go over two concrete constructions of the setting, show examples of one construction on a toy vector space and implement the other construction on the real data from the BNC. The latter is evaluated on a disambiguation task on two experiments: for intransitive verbs from \cite{Lapata} and for transitive verbs developed by us. The categorical model slightly improves the results of the first experiment and betters them in the second one. 

To draw a closer connection with the subject area of the workshop, we would like to recall  that sentences of natural language are compound systems, whose meanings exceed the meanings of their parts.  Compound systems  are a phenomena studied  by many sciences, findings  thereof should as well provide valuable insights for natural language processing. In fact, some of the above observations and previous results were led by the use of  compact categories in  compound quantum systems~\cite{AbrCoe}. The caps that connect subject and verb from afar are used to model nonlocal correlations in entangled Bell states; meanings of verbs are represented as superposed states that let the information flow between their subjects and objects and further act on it. Even on the level of single quantum systems, there are similarities to  the distributional meanings of words: both are modeled using vector spaces. Motivated by this~\cite{Rijsbergen,Widdows} have used the methods of  quantum logic to provide logical and geometric structures for information retrieval and have also obtained better results in practice.   We hope and aim to study the  modular extension of the quantum logic methods to tensor spaces of our approach. There are other  approaches to natural language processing that use compound quantum systems but  which do not focus on  distributional models, for example see~\cite{Bruza}.

Other areas of future work include  creating and running more complex experiments that involve adjectives and adverbs,  working with larger corpora such as the WaCKy, and interpreting stop words such as relative pronouns \emph{who, which}, conjunctives \emph{and, or}, and quantifiers \emph{every, some}.

\end{document}